\documentclass[letterpaper, 10 pt, conference]{ieeeconf} 
\IEEEoverridecommandlockouts                              % This command is only needed if 
\usepackage{graphics} % for pdf, bitmapped graphics files
\usepackage{epsfig} % for postscript graphics files
\usepackage{mathptmx} % assumes new font selection scheme installed
\usepackage{times} % assumes new font selection scheme installed
\usepackage{amsmath} % assumes amsmath package installed
\usepackage{amssymb}  % assumes amsmath package installed
\usepackage{algorithm}
\usepackage{cite}
\usepackage{algpseudocode}
\usepackage{multirow}
\usepackage{glossaries}
\usepackage{CJK}
\usepackage{booktabs}
\setacronymstyle{long-short}
\usepackage[dvipsnames]{xcolor}
\usepackage{hyperref}
\usepackage{enumerate}
\usepackage{microtype} % reduce spaces
\DeclareUnicodeCharacter{2212}{-}

\usepackage{tikz}
\newcommand\copyrightnotice[1]{
    \begin{tikzpicture}[remember picture,overlay]
    \node[anchor=north,xshift=-65,yshift=-12pt] at (current page.north) {\parbox{\dimexpr0.75\textwidth-\fboxsep-\fboxrule\relax}{\scriptsize #1}};
    \end{tikzpicture}
}

\title{\LARGE \bf Anticipatory Fall Detection in Humans with Hybrid Directed Graph Neural Networks and Long Short-Term Memory}

\author{Younggeol Cho, Gokhan Solak, Olivia Nocentini, \\Marta Lorenzini, Andrea Fortuna, and Arash Ajoudani% <-this % stops a space
\thanks{The authors are with the Human-Robot Interfaces and Interaction Laboratory (HRI$^2$), Istituto Italiano di Tecnologia, Genoa, Italy. Corresponding author's email: {\tt\small younggeol.cho@iit.it}}
\thanks{This work was supported in part by the European Union’s Horizon 2020 research and innovation program under Grant Agreement No. 871237 (SOPHIA) and by the BRIC LABORIUS Project.}% <-this % stops a space
}
\begin{document}
\maketitle
\thispagestyle{empty}
\pagestyle{empty}

%%%%%%%%%%%%%%%%%%%%%%%%%%%%%%%%%%%%%%%%%%%%%%%%%%%%%%%%%%%%%%%%%%%%%%%%%%%%%%%%
\begin{abstract}
Detecting and preventing falls in humans is a critical component of assistive robotic systems. While significant progress has been made in detecting falls, the prediction of falls before they happen, and analysis of the transient state between stability and an impending fall remain unexplored. In this paper, we propose a anticipatory fall detection method that utilizes a hybrid model combining Dynamic Graph Neural Networks (DGNN) with Long Short-Term Memory (LSTM) networks that decoupled the motion prediction and gait classification tasks to anticipate falls with high accuracy. Our approach employs real-time skeletal features extracted from video sequences as input for the proposed model. The DGNN acts as a classifier, distinguishing between three gait states: stable, transient, and fall. The LSTM-based network then predicts human movement in subsequent time steps, enabling early detection of falls. The proposed model was trained and validated using the OUMVLP-Pose and URFD datasets, demonstrating superior performance in terms of prediction error and recognition accuracy compared to models relying solely on DGNN and models from literature. The results indicate that decoupling prediction and classification improves performance compared to addressing the unified problem using only the DGNN. Furthermore, our method allows for the monitoring of the transient state, offering valuable insights that could enhance the functionality of advanced assistance systems.
\end{abstract}
%%%%%%%%%
\copyrightnotice{\copyright 2025 IEEE.  Personal use of this material is permitted.  Permission from IEEE must be obtained for all other uses, in any current or future media, including reprinting/republishing this material for advertising or promotional purposes, creating new collective works, for resale or redistribution to servers or lists, or reuse of any copyrighted component of this work in other works.}
%%%%%%%%%%%%%%%%%%%%%%%%%%%%%%%%%%%%%%%%%%%%%%%%%%%%%%%%%%%%%%%%%%%%%%%%%%%%%%%%
\section{INTRODUCTION}
\label{sec: introduction}
The ageing global population presents a pressing challenge in healthcare, and gait assistance systems are becoming globally crucial in supporting the elderly and patients. Particularly, detection and prevention of falls, which can cause serious injury, has been developed in recent research. To be able to detect falls, various types of sensor systems have been tested in previous research. Authors in \cite{yu2020novel} utilized the inertial sensors data, such as accelerometer and gyroscope, acquired from the public dataset to predict the pre-impact fall phase. Another paper \cite{wang2024spatio} introduced their own sensor system that can measure linear acceleration, angular speed, and 3-axis geometric information by IMU. They reconstructed a dynamic topology of lower limb motion to use for fall detection. Besides, other types of sensors such as pressure, acoustic, radar, etc. have been explored to improve fall recognition accuracy \cite{singh2020sensor}. Recent studies are more focused on utilizing vision-based sensors because wearable sensor-based systems are bulky and users are restricted in their movement. Also, vision-related deep learning technologies have been developing and they provide richer information of human movement.

The major approaches for vision processing in fall detection are silhouette-based and pose-based \cite{rastogi2022human}. Silhouette-based vision processing analyzes the outline and shape of the human. Pose-based approaches, on the other hand, focus on detecting key body joints and skeletal structures through pose estimation. These methods analyze the position and movement of body parts, such as the head, torso, and limbs, to determine whether the body posture corresponds to a fall. This paper focuses on the pose-based approach that has advantages for the real-time controlled assistive robot system. It provides more detailed and precise information of the human's posture than the silhouette. Moreover, real-time availability and robustness are the most crucial elements of assistive robot systems. For these reasons, recent papers are utilizing pose detection and deep learning techniques for detecting falls.
Authors in \cite{ramirez2021fall} achieved high recognition accuracy of fall and daily activities by using skeleton features and various linear and machine learning-based models. Another group \cite{lin2020framework} developed a fall detection framework using OpenPose skeleton extraction \cite{cao2017realtime} and a hybrid model of Long Short-Term Memory (LSTM) and Gated Recurrent Unit (GRU) to identify joint movements, achieving high accuracy without the need for wearable sensors. Authors in \cite{bharathi2024real} proposed an attention-based LSTM model that improved recognition accuracy of various human actions including falls using skeletal data. In another research \cite{xu2020fall}, Convolutional Neural Network (CNN) and extracted skeletal pose were used for fall recognition and showed high accuracy. In this context, many research groups have developed fall recognition methods based on machine learning techniques, and they showed high accuracy in various datasets \cite{rastogi2022human}. Among them, Directed graph neural network (DGNN) \cite{shi2019skeleton} showed superior performances in human action recognition, including fall detection, by utilizing the spatial and temporal skeleton features. This was achieved by modeling the kinematic dependency between the joints and bones, and their relationships through a directed acyclic graph, capturing both local and global dependencies for more accurate action recognition.

As mentioned earlier, fall detection algorithms have been developed in recent research, demonstrating high accuracy and real-time performance. However, recognizing only the current state of a fall is insufficient for effective control of assistive robots. Anticipating future human movements is essential for timely intervention and fall prevention. We may find a solution from recent studies that have proposed various deep learning-based models for movement anticipation to use for proactive fall detection. One research group proposed a scalable recurrent neural network (RNN) architecture for human motion prediction, demonstrating superior performance in short-term predictions \cite{martinez2017human}. Another group introduced an encoding-decoding deep network capable of learning a generic representation of human motion, enabling the prediction of near-future movements, which can be applied to human action anticipation \cite{butepage2017deep}. Additionally, a recent paper \cite{nouisser2022deep} presented a Convolutional LSTM-based network for anticipating future movements and recognizing falls using a random forest classification method. Rather than focusing on movement anticipation, Authors in \cite{zhao2024mss} proposed a fall risk prediction algorithm that estimates fall risk based on current skeletal features, which can be used for early fall prevention. Although significant progress has been made in fall detection and movement anticipation, further development is needed to enhance fall anticipation performance. In particular, analyzing the transient state between stable walking and falling is crucial for enabling assistive devices to respond in advance with high accuracy. Current research either overlooks the duration of this transient state or only classifies movements into two categories: fall and non-fall.

This study presents our first but significant steps towards fall anticipation for an assistive walking robot. We introduced the Walking Assistive Omni-Directional Exo-Robot (WANDER) and its control strategy in an earlier work \cite{fortuna2024personalizable}. This research is essential for implementing fall anticipation and prevention functions to enhance safety. In this paper, our aim is to design a hybrid neural network that combines DGNN and LSTM, decoupling the networks to handle motion prediction and gait classification tasks separately. The DGNN architecture was optimized for accurate fall recognition, while the LSTM-based human movement anticipation network was developed to predict future skeletal pose features, which serve as inputs to the DGNN to ultimately recognize the fall states before they happen. We utilized the UR Fall and OUMVLP-Pose datasets to train and validate the proposed network. The results demonstrated that using separate networks for fall recognition and future pose anticipation improved overall performance. Furthermore, we showed that predicting future skeletal pose features can help analyze movement trends, which can be leveraged by assistive robots to intervene in pre-fall or recovery situations.

%%%%%%%%%%%%%%%%%%%%%%%%%%%%%%%%%%%%%%%%%%%%%%%%%%%%
\section{Methodology}
\label{sec: methodology}

\subsection{Model Overview}
\label{sec: overview}
\begin{figure}[t]
\vspace{0.3cm}
   \centering
    \centerline{\includegraphics[width=1.0\columnwidth]{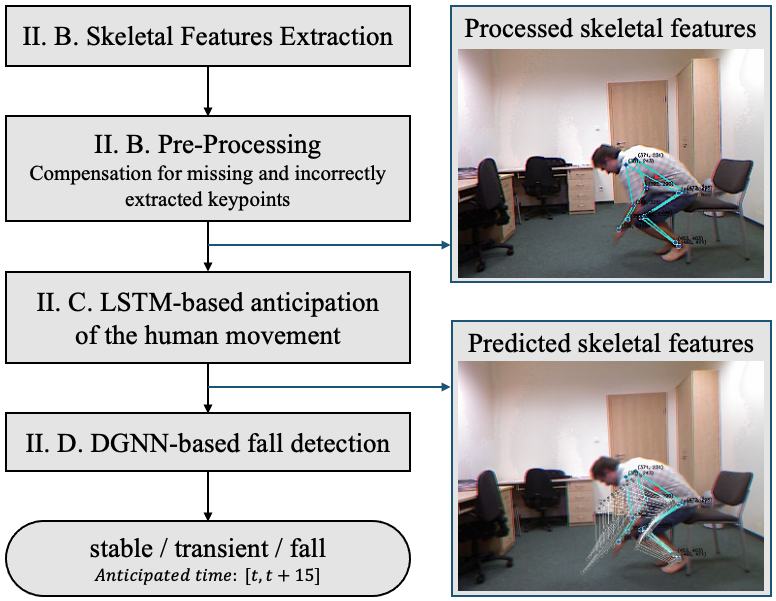}}
 \caption{Overview of the proposed fall anticipation model. It begins with the skeletal feature extraction and processing. The proposed model primarily consists of decoupled neural networks designed for movement anticipation and gait state recognition.}
\label{fig:overview}
\end{figure}

The proposed architecture for fall anticipation begins with the extraction of skeletal features from video frames (see Fig. \ref{fig:overview}). These extracted features are then processed to compensate for missing or incorrectly extracted points by considering the human body's configuration (Section II.B). The processed skeletal features are passed through an LSTM-based network to predict human movement from the current to a future state (Section II.C). The predicted skeletal features are subsequently fed into a DGNN-based fall detection network, which classifies the human state into three categories: stable, transient, and fall (Section II.D). 
%%%%%%%%%%%%%%
%% Figure 1 %%
%%%%%%%%%%%%%%
%%%%%%%%%%%%%%
%% Figure 2 %%
%%%%%%%%%%%%%%
\begin{figure}[b]
\vspace{0.3cm}
   \centering
    \centerline{\includegraphics[width=0.8\columnwidth]{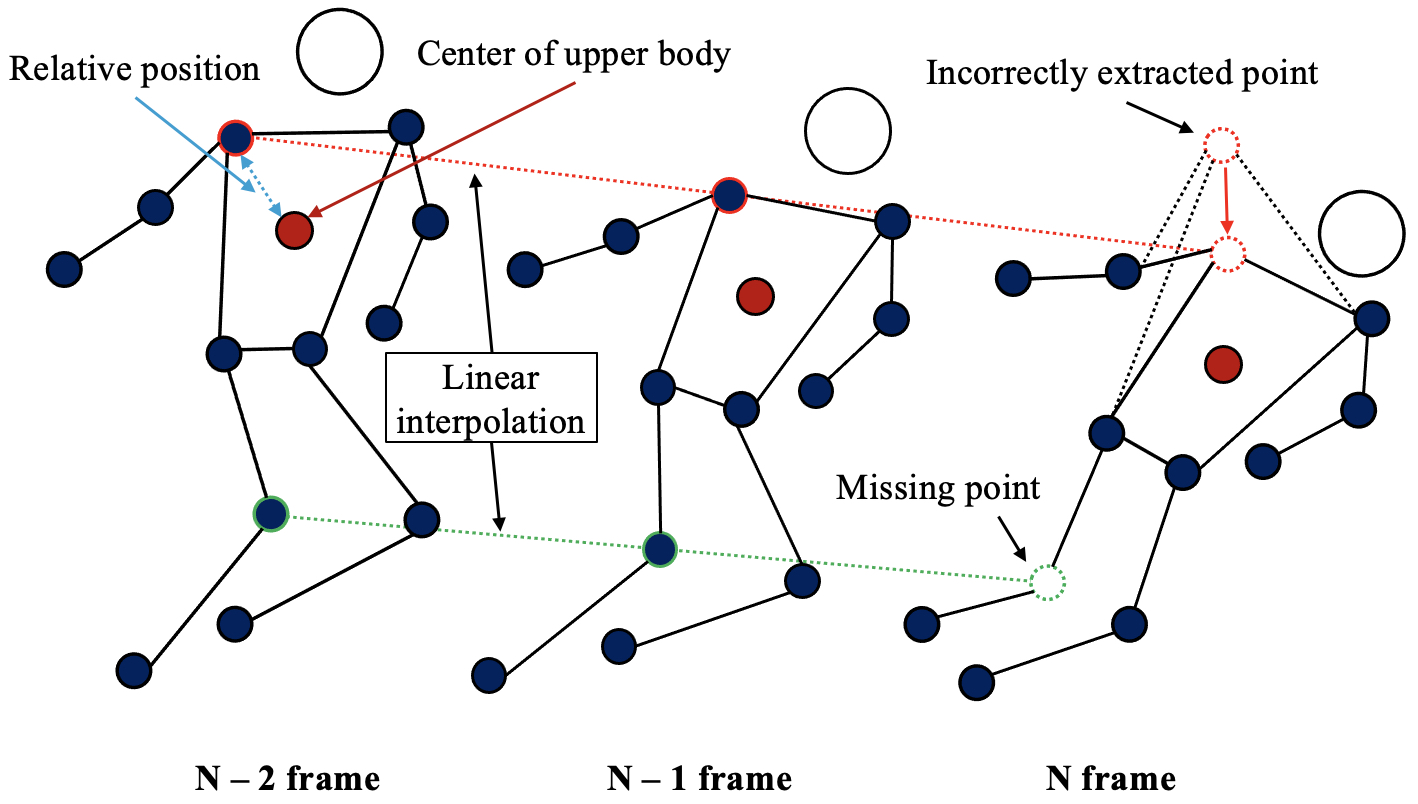}}
 \caption{Pre-processing of extracted skeletal features. The missing or incorrect keypoints were linearly compensated by using past frames. The relative positions of the keypoints were determined by calculating the Euclidean distance from the center of the upper body.}
\label{fig:skeleton-frames}
\end{figure}
\subsection{Skeletal Pose Extraction and Processing}
\label{sec: skeletal pose}
The skeletal pose features such as keypoints of the body (e.g., joints) were extracted by a recent pose estimation model, YOLOv8-Pose \cite{Jocher2023}. We extracted 12 keypoints from the body and processed them to derive robust features for analysis. In some cases, the pose estimation model encountered challenges, leading to missing or incorrectly extracted keypoints. To address this, we implemented preprocessing techniques to handle these inaccuracies and ensure the robustness of the feature set. First, missing keypoints were compensated using linear interpolation based on the previous two points (see Fig.~\ref{fig:skeleton-frames}. green circle). Given that the video frame rate is significantly higher than the natural movement bandwidth of the human body, linear interpolation provides a sufficient and reliable method for restoring these missing points. Second, we corrected the keypoints that were extracted incorrectly. The pose estimation model occasionally places keypoints incorrectly outside the body's expected range. We identified these errors by observing that the keypoints moved at an unrealistic speed, far exceeding the natural range of human movement. Through empirical testing, we established a threshold value. If the change in a keypoint's location between frames exceeds this threshold, we correct the error using linear interpolation (see Fig.~\ref{fig:skeleton-frames}. red circle). As the final step, we applied a second-order Butterworth low-pass filter with a cutoff frequency of 10 Hz to eliminate high-frequency noise in the keypoint changes, which is inherently introduced by the pose estimation model.

\subsection{Anticipation of Human Movement}
\label{sec: prediction}
We utilized a multi-layer LSTM neural network capable of processing sequential data from extracted keypoints over time. The input features for this network were calculated as the relative positions from the center of the upper body, which was calculated using keypoints from the shoulder and pelvis keypoints. This approach makes the anticipation results more robust to variations in the human’s location within the video frame. The network architecture includes LSTM layers, followed by a dropout layer to prevent overfitting. A fully connected layer follows, producing the same number of nodes as the input layer, corresponding to the relative keypoints. The network weights were initialized using Xavier normalization. The parameter details were presented in the experimental details section.

%%%%%%%%%%%%%%%%%%%%%%%%%%%%%%%%%%%%%%%%%%%%%%%%%%%%%%%%%%%%%%%%%%%%%%
\subsection{Fall Detection} % DGNN ------- Thanks my friend, Gokhan :)
\label{sec: detection}
We use the DGNN \cite{shi2019skeleton}, a state-of-the-art graph-based classification method to recognise gait state (stable, transient, fall). 
This method directly uses the skeleton information without any feature engineering. 
A skeleton is defined by the keypoints (as extracted in Sec.~\ref{sec: skeletal pose}) and the predefined edges between them, e.g., the wrist is connected to the elbow. 
The graph neural network can exploit the edge structure of a skeleton to propagate information through the network, increasing the influence of directly connected keypoints on each other.
The DGNN assumes directed edges, thus preserving the parent-child hierarchy of the graph nodes, and uses the bone information, that is, relative position of the keypoints with their neighbors, to obtain better classification results. 
Finally, the DGNN adapts the graph structure (adjacency matrix) during training to give more attention to the relationship between more related keypoints. 

In addition to the spatial information through the graph structure, the DGNN also exploits the temporal information in an action sequence by temporal convolution.
In the source paper \cite{shi2019skeleton}, each action sequence belongs to a single action class and the full sequence (of variable size) is fed to the classifier. 
However, in our task, knowing the exact moment of the fall is crucial. 
Therefore, we feed the data as a time window of past $T$ frames (window size) to classify the current frame. 
So, in our method, the DGNN classifies each time frame instantly. 

In this paper, we also explore the idea of prediction using the DGNN only, and use it as our baseline in the experiments. The \textit{DGNN-only} approach is based on training the DGNN by giving the future fall states as the labels, instead of the current ones. In other words, we add a time-offset to the target class, so that the model is conditioned for prediction. 

%%%%%%%%%%%%%%%%%%%%%%%%%%%%%%%%%%%%%%%%%%%%%%%%%%%%%%%%%%%%%%%%%%%%%%
\subsection{Transient State Analysis}
\label{sec: transient}
To analyze the trends in state transitions, particularly within the transient state, we employed Principal Component Analysis (PCA) to reduce the high-dimensional feature space into a two-dimensional representation. This dimensionality reduction enabled us to visualize distinct class clusters (stable, transient, and fall) and observe the predicted trajectories as they evolved over time. By mapping the temporal sequence of pose keypoints, we gained insight into the progression of instances as they transition from a stable state through the transient phase, ultimately to a fall state. 
%%%%%%%%%%%%%%%%%%%%%%%%%%%%%%%%%%%%%%%%%%%%%%%%%%%%
%%%%%%%%%%%%%%
%% Figure 3 %%
%%%%%%%%%%%%%%
\begin{figure*}[ht]
\vspace{0.3cm}
   \centering
    \centerline{\includegraphics[width=2.0\columnwidth]{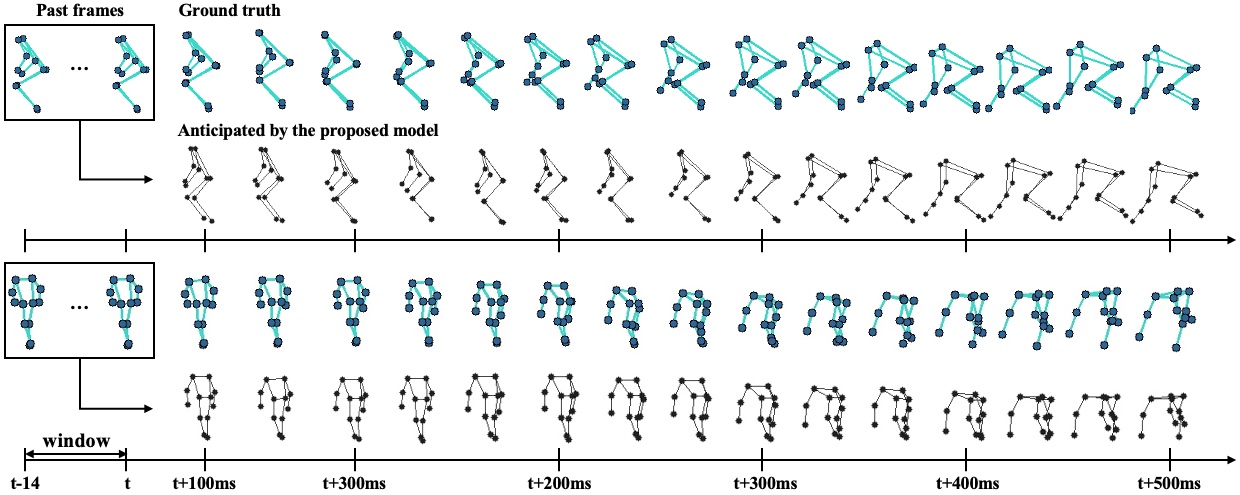}}
 \caption{A sequence prediction example for skeletal motion. The first and third rows shows the ground truth, representing the actual movements over time. The second and third row shows the anticipated motion as predicted by a proposed model, starting from the current time step (t) and projecting forward. A sliding window (spanning from t-14 to t) provides the model with past frames to predict future movement.}
\label{fig:prediction-example}
\end{figure*}

\section{Experimental Details}
\label{sec: experimental}

\subsection{Datasets}
\label{sec: datasets}

\begin{table}[b]
\setlength{\tabcolsep}{18pt}
\caption{Network structures and their parameters.}
\begin{tabular}{l|l}
\hline\hline
\multicolumn{2}{c}{LSTM-based movement anticipation} \\
\hline
Parameters             & Value                         \\
\hline
Input                  & 24 (keypoints x, y)           \\
Hidden layers          & 512                           \\
LSTM layers            & 2                             \\
Fully connected layers & {[}512, 24{]}                 \\
Output                 & 24 (keypoints x, y)           \\
Dropout probability    & 0.5                           \\
Window size            & 15                            \\
Loss function          & Cross Entropy                 \\
\hline\hline
\multicolumn{2}{c}{DGNN-based fall detection}        \\
\hline
Parameters             & Value                         \\
\hline
Input                  & 24 (keypoints x, y), 12 links \\
1st DGNN layer         & {[}12, 16{]}                  \\
2nd DGNN layer         & {[}16, 32{]}                  \\
3rd DGNN layer         & {[}32, 64{]}                 \\
Dropout probability    & 0.5                           \\
Fully connected layer  & {[}128, 3{]}                  \\
Output                 & 3 (stable/transient/fall)     \\
Window size            & 15                            \\
Loss function          & Smooth L1 Loss                 \\
\hline
\end{tabular}
\end{table}

To train the DGNN-based fall recognition network, we utilized the UR Fall Detection (URFD) dataset \cite{kwolek2014human}, which contains sequences of human activities such as falls and routine movements like walking, standing, and sitting. The data was collected through various sensors, including RGB cameras and accelerometers. We extracted skeletal features from RGB images, which served as input for the fall recognition network. For predicting future movement, we used both the URFD and OUMVLP-Pose datasets \cite{an2020performance}, a subset of the OU-Mult View Large Pose (OU-MVLP) dataset, which provides human skeletal features during walking. The inclusion of multi-view images helps improve anticipation accuracy and robustness across different camera perspectives.

\subsection{Network parameters and Training Conditions}
\label{sec: training}
The structures and parameters of the two network architectures for movement anticipation and fall detection are presented in the Table 1. The LSTM-based model is designed to anticipate future movement, incorporating LSTM layers, fully connected layers, and dropout regularization. This model was pre-trained using the OUMVLP-Pose dataset for normal walking anticipation and fine-tuned with the URFD dataset to anticipate motion during transient and fall states. The DGNN-based model focuses on fall detection, processing both keypoints and links (bones), with multiple DGNN layers, fully connected layers, and a softmax function to classify three states: stable, transient, and fall. The DGNN-based model was trained using the URFD dataset primarily to balance the dataset size. In training section for both models, We divided the training and test data into 80$\%$ and 20$\%$, respectively. We utilized the Adam optimizer \cite{kingma2014adam} and empirically selected an initial learning rate of 0.001. The learning rate was decreased every 16 epochs, with a maximum of 300 epochs set for training.

\subsection{Evaluations}
\label{sec: evaluations} 
The movement anticipation performance was evaluated using the Normalized Euclidean distance ($\bar{d}$). between the ground truth and anticipated relative keypoints.
\begin{equation}
\label{eq1}
\resizebox{.2\textwidth}{!}{
$\bar{d} = \frac{1}{m} \sum_{i=1}^{m} \|\mathbf{p_a}_i - \mathbf{p_t}_{i+t_a}\|_2$}
\end{equation}
$p_a$ and $p_t$ represent the anticipated and ground truth keypoints, respectively. The distance was averaged across all time frames, with $t_a$ denoting the anticipated time step. For qualitative analysis, we presented both ground truth and anticipated human skeletal features.

The fall anticipation performance was evaluated based on fall recognition accuracy concerning the anticipated time step, ranging from 0 to 500 milliseconds. The four models from literature was compared with the proposed model. 

The transient state was qualitatively analyzed using Principal Component Analysis (PCA) to visualize the input features onto the PCA plane. The trajectory of the PCA features was then plotted to analyze the trend across the current and future time steps.

%%%%%%%%%%%%%%%%%%%%%%%%%%%%%%%%%%%%%%%%%%%%%%%%%%%%
\section{Result}
\label{sec: result}
%%%%%%%%%%%%%%
%% Figure 4 %%
%%%%%%%%%%%%%%
\begin{table}[b]
\caption{Anticipation error (Euclidean distance).}
\begin{tabular}{c|cccccc}
\hline\hline
\multirow{2}{*}{}  & \multicolumn{6}{c}{Anticipation time {[}s{]}}                                                                                                      \\ \cline{2-7} 
                   & \multicolumn{1}{c|}{0} & \multicolumn{1}{c|}{0.1}   & \multicolumn{1}{c|}{0.2}   & \multicolumn{1}{c|}{0.3}    & \multicolumn{1}{c|}{0.4}   & 0,5   \\ \hline
Average            & \multicolumn{1}{c|}{0} & \multicolumn{1}{c|}{0.031} & \multicolumn{1}{c|}{0.031} & \multicolumn{1}{c|}{0.0327} & \multicolumn{1}{c|}{0.029} & 0.034 \\ \hline
standard deviation & \multicolumn{1}{c|}{0} & \multicolumn{1}{c|}{0.010} & \multicolumn{1}{c|}{0.006} & \multicolumn{1}{c|}{0.005}  & \multicolumn{1}{c|}{0.007} & 0.007 \\ \hline\hline
\end{tabular}
\label{fig:anticipation_error}
\end{table}
\begin{table*}[h]
\caption{Comparison of the fall anticipation accuracy}
\setlength{\tabcolsep}{12pt}
\begin{tabular}{l|c|cccccc}
\hline\hline
 & \multicolumn{1}{l|}{} & \multicolumn{6}{c|}{Anticipating Time step (UR Fall Dataset)} \\ \hline\hline
Base Model & \begin{tabular}[c]{@{}c@{}}Num. of \\    \\ classes\end{tabular} & \multicolumn{1}{c|}{0} & \multicolumn{1}{c|}{3 (100ms)} & \multicolumn{1}{c|}{6 (200ms)} & \multicolumn{1}{c|}{9 (300ms)} & \multicolumn{1}{c|}{12 (400ms)} & 15 (500ms) \\ \hline
DGNN only & 3 & \multicolumn{1}{c|}{0.958} & \multicolumn{1}{c|}{0.912} & \multicolumn{1}{c|}{0.908} & \multicolumn{1}{c|}{0.814} & \multicolumn{1}{c|}{0.785} & 0.764 \\ \hline
CNN \cite{xu2020fall} & 2 & \multicolumn{1}{c|}{0.917} & \multicolumn{1}{c|}{-} & \multicolumn{1}{c|}{-} & \multicolumn{1}{c|}{-} & \multicolumn{1}{c|}{-} & - \\ \hline
VGG16 Net \cite{chhetri2021deep} & 2 & \multicolumn{1}{c|}{0.951} & \multicolumn{1}{c|}{-} & \multicolumn{1}{c|}{-} & \multicolumn{1}{c|}{-} & \multicolumn{1}{c|}{-} & - \\ \hline
RNN \cite{theodoridis2018human} & 2 & \multicolumn{1}{c|}{0.928} & \multicolumn{1}{c|}{-} & \multicolumn{1}{c|}{-} & \multicolumn{1}{c|}{-} & \multicolumn{1}{c|}{-} & - \\ \hline
ConvAutoencoder \cite{cai2020vision} & 2 & \multicolumn{1}{c|}{0.962} & \multicolumn{1}{c|}{-} & \multicolumn{1}{c|}{-} & \multicolumn{1}{c|}{-} & \multicolumn{1}{c|}{-} & - \\ \hline
LSTM+DGNN (proposed) & 3 & \multicolumn{1}{c|}{0.958} & \multicolumn{1}{c|}{0.912} & \multicolumn{1}{c|}{0.886} & \multicolumn{1}{c|}{0.900} & \multicolumn{1}{c|}{0.888} & 0.894 \\ \hline
\end{tabular}
\label{fig:Comparison}
\end{table*}
\subsection{Movement Anticipation Performance}
\label{sec: result_movement}
\begin{figure}[t]
   \centering
    \centerline{\includegraphics[width=1.0\columnwidth]{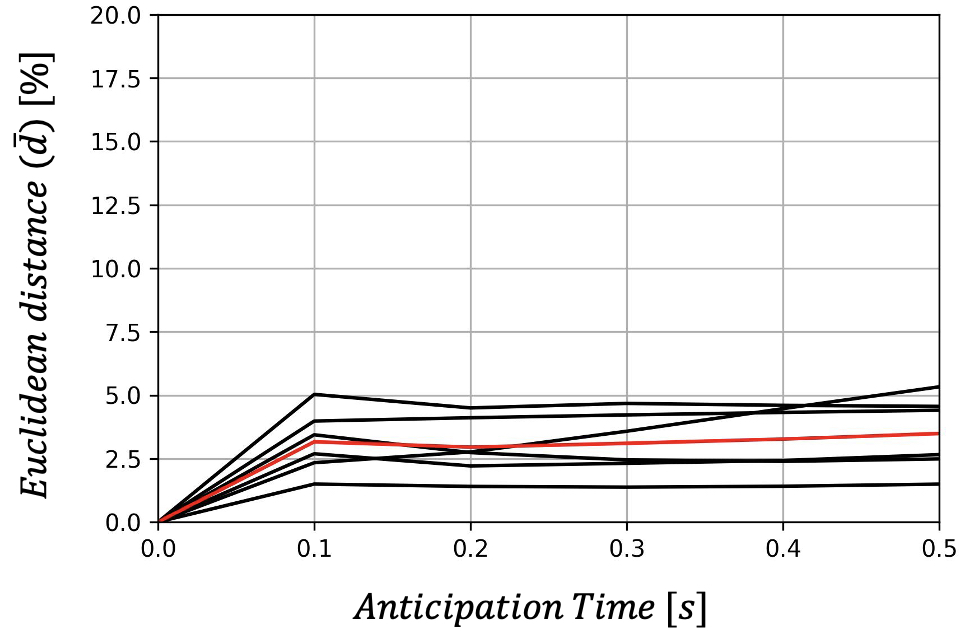}}
 \caption{Anticipation error. The red line represent the average of anticipation errors from test sets.}
\label{fig:Anticipation error}
\end{figure}

Sequence anticipation examples are presented in Fig.~\ref{fig:prediction-example}. By utilizing features from previous frames, the proposed model is able to anticipate up to 500 ms into the future. The anticipated skeletal features (shown in black in Fig.~\ref{fig:prediction-example}) exhibit a configuration similar to the ground truth (shown in cyan in Fig.~\ref{fig:prediction-example}). The Euclidean distance, used to assess movement anticipation performance, was presented in relation to the anticipation time step, with measurements taken at 0.1-second intervals (see Fig.~\ref{fig:Anticipation error} , Table.~\ref{fig:anticipation_error}) for quantitative analysis. Detailed values are provided in Table 2. The results reveal that the average normalized Euclidean distance across all anticipation times was approximately 0.03 (3$\%$), with a standard deviation of less than 0.01. Notably, the proposed model demonstrates the ability to predict movement 500 milliseconds into the future, achieving an average Euclidean distance of 0.034 with a standard deviation of 0.007. the model's ability to anticipate future movements reaches an average Euclidean distance of 0.031 with a standard deviation of 0.007, suggesting that the model maintains robust performance even as the anticipation window increases up to 0.5$s$.

\subsection{Fall Anticipation Performance}
\label{sec: result_recognition}
The fall anticipation performance was evaluated by comparing various neural network-based models on the URFD dataset. Table ~\ref{fig:Comparison} presents the fall anticipation accuracy at different time steps, ranging from 0 to 500 milliseconds. The comparative models include DGNN, CNN, VGG16 Net, RNN, and ConvAutoencoder. All models, except DGNN, recognize the fall state at the current time only, without anticipating future movements. Therefore, the proposed model was compared with other models at the current time step only. The proposed model achieves a fall anticipation accuracy of 0.958, demonstrating competitive performance compared to other recent models. While the DGNN model's accuracy decreases from 0.958 at 0 ms to 0.764 at 500 ms, the proposed LSTM+DGNN model maintains strong performance, starting at 0.958 at 0 ms and achieving 0.894 at 500 ms, indicating superior overall anticipation accuracy. 

\subsection{Transient analysis}
\label{sec: result_transient}
\begin{figure}[!h]
\vspace{0.3cm}
   \centering
    \centerline{\includegraphics[width=1.0\columnwidth]{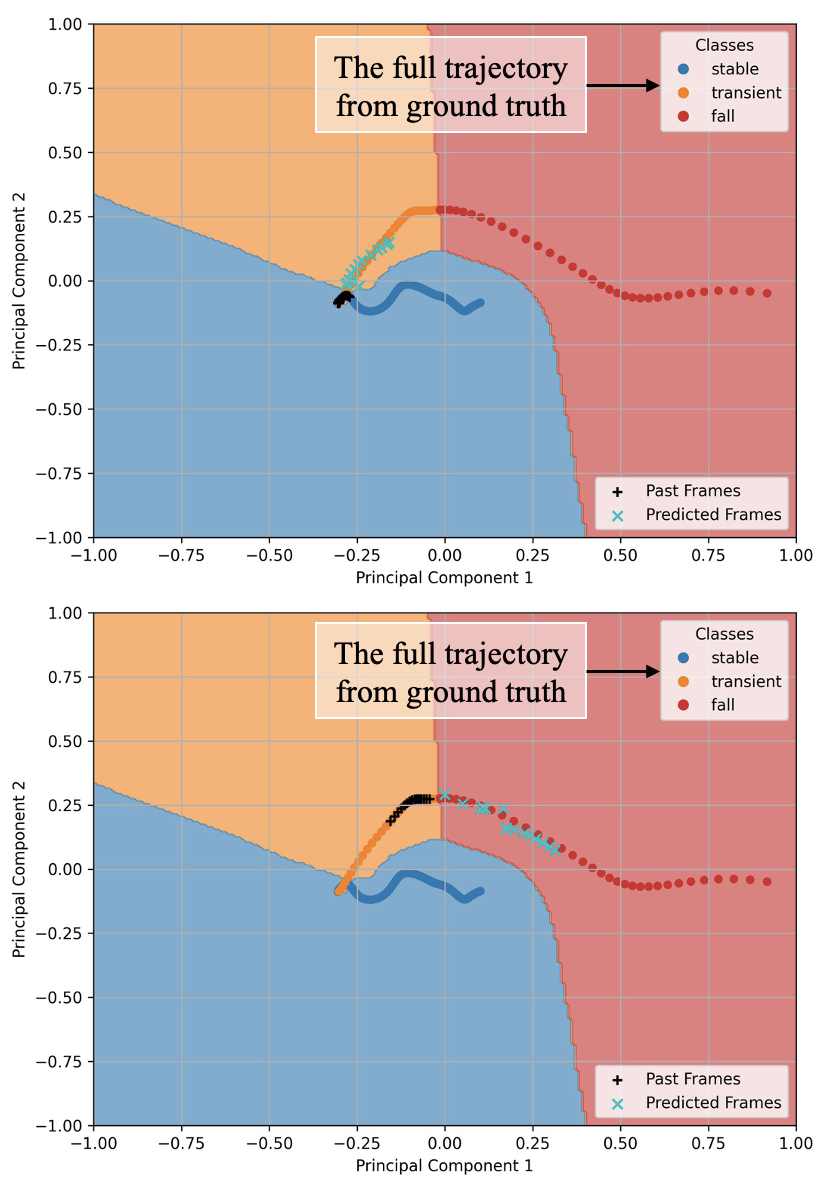}}
 \caption{The transient analysis using Principal Component Analysis (PCA) plots classifies gait into three categories: stable (blue), transient (orange), and fall (red). Current states are shown in black, while anticipated states are in cyan. The trends near the transient state can be interpreted with richer information, such as the location and speed of the projected feature points.}
\label{fig:transient-analysis}
\end{figure}

We presented input features and the results of a classification model predicting three different classes: ``stable," ``transient," and ``fall," into 2D plane based on the PCA projection. A critical aspect of the analysis is when the projected features cross the decision boundary. 
For example, as shown in Fig.~\ref{fig:transient-analysis}, the features of past frames (black markers) are near the boundary, allowing us to anticipate that the features of future frames (cyan markers) will move into a different class area. Based on the trajectory of these markers, we can assess the current state by determining how close the features are to the boundary and how quickly they may cross into the fall state area. 
In the graph shown in Fig.~\ref{fig:transient-analysis}, we interpret that the current state is nearing the boundary between transient and fall states. As the projected features move deeper into the fall state area, we can anticipate a fall, indicating that the person will likely need assistance to regain balance, possibly with the help of an assistive device. Anticipating the fall state alone is not sufficient to effectively assist a person. However, the analysis of the projected feature points and their trajectories offers richer information, such as direction, speed, and position within the class area. This information can be highly valuable for future assistive control algorithms, enabling more proactive and precise interventions.
%%%%%%%%%%%%%%
%% Figure 5 %%
%%%%%%%%%%%%%%
%%%%%%%%%%%%%%%%%%%%%%%%%%%%%%%%%%%%%%%%%%%%%%%%%%%%
\section{Discussion}
\label{sec: discussion}

The results demonstrated that our proposed LSTM-DGNN based model anticipated future human movement with low error rates and accurately predicted the future fall state up to 500 milliseconds in advance, achieving high accuracy. In addition, because this paper represents preliminary research in the field of walking assistance, where the analysis of the transient state is particularly important. Understanding the transient state and its dynamics is crucial for developing effective assistance strategies The transient analysis in this paper was qualitative analysis that can give the insight to interpret the trend of the time step ahead. In future works, by analyzing the transient state base on PCA dimensional reduction projection, we can compute the distance between the current point and the center of each class for quantitative analysis. The center of each class can be calculated as the center of the corresponding points in the reduced feature space, determined by averaging the coordinates of all points belonging to a specific class. The distance between a current projected feature point and the class center can be useful as a information for assistive device. Additionally, from the variation in these distances over time, we can calculate the dynamics of the transition, including the rate and direction of change. Transient analysis will be crucial for optimizing the control strategies of assistive robots, enabling them to anticipate and respond to transitions between states more effectively. In future work, these insights can be used to improve real-time human support, enhancing safety and stability during critical moments such as potential falls.

This paper has several limitations that will be addressed in future work. We utilized the URFD and OUMVLP-Pose datasets to train and evaluate the proposed model. However, the URFD dataset contains only 30 fall cases, which is insufficient for generalizing the model to diverse real-world scenarios. In future research, we plan to incorporate additional fall-related datasets and create our own dataset to fine-tune the model for assistive applications. Moreover, the movement anticipation model is currently based on a simple LSTM network, which can be improved by exploring more advanced deep learning architectures \cite{nouisser2022deep, rastogi2021systematic}. In this paper, other models can be used for anticipating movement, but we test the idea of dividing of anticipating movement and fall detection.

The proposed model was compared with models that only detect the current state, including falls. Research on human movement anticipation and fall detection has typically been conducted separately, and combining these to anticipate falls in advance has not yet been thoroughly explored. As a result, a direct performance comparison with existing models was difficult to conduct and should be addressed in future work.

%%%%%%%%%%%%%%%%%%%%%%%%%%%%%%%%%%%%%%%%%%%%%%%%%%%%
\section{Conclusion}
\label{sec: conclusion}

In this preliminary study, we explored the problem of fall prediction which can be divided it into two parts: anticipating the future motion states and recognizing the fall states for a given state. We designed our method to deal with these sub-problems separately, answering the former using a LSTM predictor and the later using a DGNN classifier. We evaluated our method against the common approach of using a single model for the whole problem. Our experiment results show that our method significantly increases the prediction performance over the single model approach on the publicly available URFD dataset. A separate anticipation model performs better both because it can leverage information related to gait movement better and it can make use of more general public datasets that are not focused on the fall recognition. In addition, we analyzed the transient state by visualizing the projected features in the PCA plane, which provides insight into the trend of state changes. This information can be valuable for developing assistance systems. Furthermore, we see that our results compare favorably to the previously published results on the URFD dataset, even though the previous results answer only the recognition problem. Thanks to its anticipation capability, our method can be used in a robotic gait assistance system to prevent the accidents of the elderly and patients. Consequently, our future work will focus on developing the predictive control method of such system and evaluate the practical benefit of our method.

\addtolength{\textheight}{-14.5cm} 

\bibliographystyle{IEEEtran}
\bibliography{references.bib} % automatically inserted and ordered with this command 

\end{document}